\documentclass[10pt,twocolumn,letterpaper]{article}

 \usepackage{cvpr}              % To produce the CAMERA-READY version
\usepackage[dvipsnames]{xcolor}

\definecolor{cvprblue}{rgb}{0.21,0.49,0.74}
\usepackage[pagebackref,breaklinks,colorlinks,citecolor=cvprblue]{hyperref}

\usepackage{circledsteps}

\def\paperID{15} % *** Enter the Paper ID here
\def\confName{CVPR}
\def\confYear{2024}

\title{Exploring Explainability in Video Action Recognition}

\author{Avinab Saha$^*$, Shashank Gupta$^*$, Sravan Kumar Ankireddy$^*$, Karl Chahine, Joydeep Ghosh \\
The University of Texas at Austin 
}

\begin{document}
\maketitle
\def\thefootnote{*}\footnotetext{Equal Technical Contribution. Correspondence to Avinab Saha, Shashank Gupta and Sravan Kumar Ankireddy.  Email: \{avinab.saha,shashank.gupta,sravan.ankireddy\}@utexas.edu}
\begin{abstract}
Image Classification and Video Action Recognition are perhaps the two most foundational tasks in computer vision. Consequently, explaining the inner workings of trained deep neural networks is of prime importance. While numerous efforts focus on explaining the decisions of trained deep neural networks in image classification, exploration in the domain of its temporal version, video action recognition, has been scant. In this work, we take a deeper look at this problem. We begin by revisiting Grad-CAM, one of the popular feature attribution methods for Image Classification, and its extension to Video Action Recognition tasks and examine the method's limitations. To address these, we introduce Video-TCAV, by building on TCAV for Image Classification tasks, which aims to quantify the importance of specific concepts in the decision-making process of Video Action Recognition models. As the scalable generation of concepts is still an open problem, we propose a machine-assisted approach to generate spatial and spatiotemporal concepts relevant to Video Action Recognition for testing Video-TCAV. We then establish the importance of temporally-varying concepts by demonstrating the superiority of dynamic spatiotemporal concepts over trivial spatial concepts. In conclusion, we introduce a framework for investigating hypotheses in action recognition and quantitatively testing them, thus advancing research in the explainability of deep neural networks used in video action recognition.
\end{abstract}    
\section{Introduction}
\label{sec:intro}
\indent Understanding human actions in videos is crucial for various applications like behavior analysis, video retrieval, and human-robot interaction. Human action understanding involves
recognizing, localizing, and predicting human behaviors.
The task to recognize human actions in a video is termed as \emph{Video Action Recognition}. In recent years, significant research efforts have focused on developing effective models for this task, including I3D \cite{Carreira_2017_CVPR}, SlowFast \cite{feichtenhofer2019slowfast}, and Video Swin Transformer \cite{https://doi.org/10.48550/arxiv.2106.13230}. Consequently, an intriguing avenue for further exploration lies in understanding the decision-making processes of these networks. \\
\indent Recently, multiple works have focused on analyzing decisions made by neural networks in the context of image classification tasks by developing feature attribution methods: Integrated gradients \cite{sundararajan2017axiomatic}, Class Activation Mapping (CAM) \cite{zhou2016learning}, and Grad-CAM \cite{selvaraju2017grad}. A widely embraced alternative for studying the explainability of deep neural networks in image classification tasks over traditional feature attribution methods is TCAV \cite{kim2018interpretability}. TCAV is a global explanation method focusing on more abstract details instead of granular, pixel-level changes typically associated with feature attribution methods. Although there has been significant progress in post-training explainability methods for deep learning-based image classification, very little research has been done on the applicability of these methods in the context of video action recognition. In this work, we aim to investigate this direction by exploring a feature attribution method for video action recognition, discussing its limitations, and further introducing a video counterpart to TCAV, which we refer to as Video-TCAV. Specifically, we opt for \emph{playing tennis} class from the Kinetics-400 dataset \cite{DBLP:journals/corr/KayCSZHVVGBNSZ17} to visualize the outcomes of the Grad-CAM as well as to formulate the proposed Video-TCAV framework. \\
\indent The rest of the paper is organized as follows. Section \ref{sec:featureattrib} discusses the performance of Grad-CAM when employed in the context of video action recognition to set up a baseline representing popular 
feature attribution methods. In Section \ref{sec:videotcav}, we introduce Video-TCAV, which includes an automated pipeline for generating high-level concepts and are evaluated in the proposed Video-TCAV framework. Section \ref{sec:conclude} concludes the paper by summarizing the ideas explored in this paper and discussing future research directions.

\iffalse
\begin{figure}[t]
  \centering
  \fbox{\rule{0pt}{2in} \rule{0.9\linewidth}{0pt}}
   %\includegraphics[width=0.8\linewidth]{egfigure.eps}

   \caption{Example of caption.}
   \label{fig:onecol}
\end{figure}

\begin{figure*}
  \centering
  \begin{subfigure}{0.68\linewidth}
    \fbox{\rule{0pt}{2in} \rule{.9\linewidth}{0pt}}
    \caption{An example of a subfigure.}
    \label{fig:short-a}
  \end{subfigure}
  \hfill
  \begin{subfigure}{0.28\linewidth}
    \fbox{\rule{0pt}{2in} \rule{.9\linewidth}{0pt}}
    \caption{Another example of a subfigure.}
    \label{fig:short-b}
  \end{subfigure}
  \caption{Example of a short caption, which should be centered.}
  \label{fig:short}
\end{figure*}
\fi

\section{Grad-CAM revisited}
\label{sec:featureattrib}
\iffalse

\begin{figure*}[tp]
\centering 
\begin{minipage}[b]{0.22\linewidth}
  \centering
  \subfloat[Positive Illustration]{%
    \includegraphics[clip,width=\linewidth]{figs/tennis_1.png}
    \label{fig:gradcam_tennis_1}
  }
  
\end{minipage}
\hspace{0.02\linewidth} % Adjust the horizontal space between the figures
\begin{minipage}[b]{0.22\linewidth}
  \centering
  \subfloat[Positive Illustration]{%
    \includegraphics[clip,width=\linewidth]{figs/tennis_2.png}
    \label{fig:gradcam_tennis_2}
  }
  
\end{minipage}
\hspace{0.02\linewidth}
\begin{minipage}[b]{0.22\linewidth}
  \centering
  \subfloat[Limitation]{%
    \includegraphics[clip,width=\linewidth]{figs/tennis_3.png}
    \label{fig:gradcam_limitation_tennis}
  }
\end{minipage}
\captionsetup{font=small}
\caption{Outputs of video Grad-CAM with respect to the class \textit{playing tennis} for Video Swin Transformer model. (a), (b): Grad-CAM correctly highlights the regions of movement for tennis rackets. (c): Grad-CAM focuses on the tennis court in the background and ignores the players in the frame.}
\label{fig:gradcam_illustration}
\vspace{-1.5em}
\end{figure*}
\fi

\begin{figure}[tp]
\centering 
\begin{minipage}[b]{0.37\linewidth}
  \centering
  \subfloat[Positive Illustration]{%
    \includegraphics[clip,width=\linewidth]{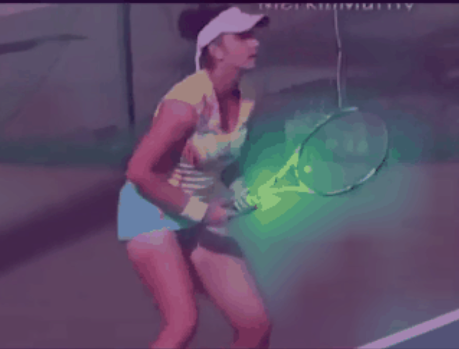}
    \label{fig:gradcam_tennis_1}
  }
  
\end{minipage}
\hspace{0.02\linewidth} % Adjust the horizontal space between the figures
\begin{minipage}[b]{0.37\linewidth}
  \centering
  \subfloat[Limitation]{%
    \includegraphics[clip,width=\linewidth]{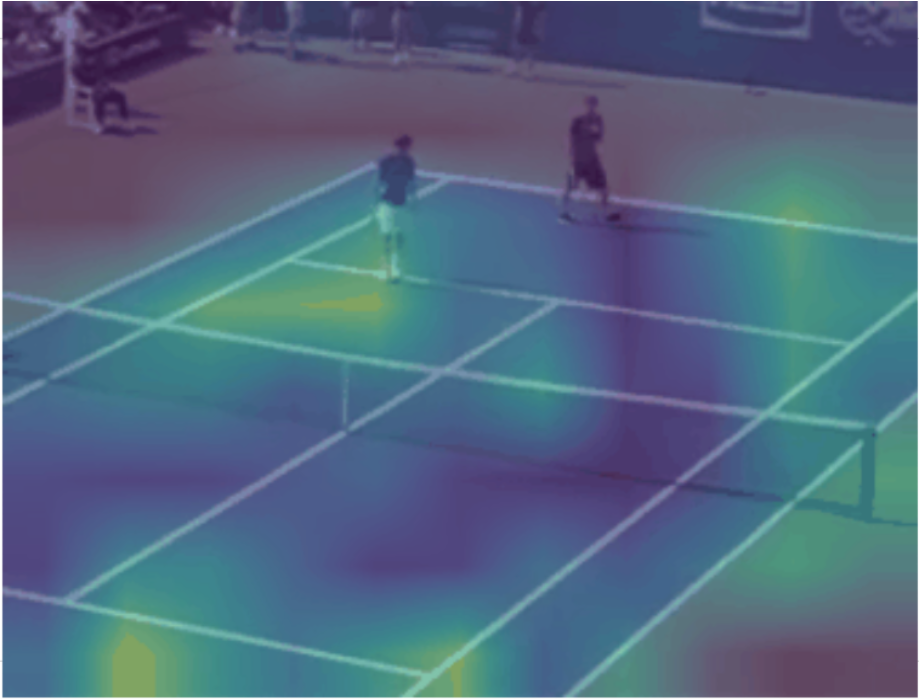}
    \label{fig:gradcam_limitation_tennis}
  }
\end{minipage}
\captionsetup{font=small}
\caption{ Grad-CAM outputs with respect to the class \textit{playing tennis} for Video Swin Transformer model. (a): Grad-CAM correctly highlights the regions of movement for tennis rackets. (b): Grad-CAM focuses on the tennis court in the background and ignores the players in the frame.}
\label{fig:gradcam_illustration}
\vspace{-1.5em}
\end{figure}
\indent 
% In our exploration of explainability for video action recognition models, 
We first revisit Gradient-weighted Class Activation Mapping (Grad-CAM)\cite{selvaraju2017grad}, a popular feature attribution method renowned for its effectiveness in understanding image recognition models. %Using Grad-CAM as a building block, efforts were made in the past to image classification models.
Class Activation Maps (CAMs) visually represent the focus areas for an image recognition model during prediction by utilizing the weights of the model's final convolutional layer.
% to generate a heatmap of the input image to highlight the significant parts of the input that contribute to the model's decision-making process. 
Building on this, Grad-CAM has emerged as a popular technique for generating visual explanations for various image recognition models by leveraging the gradient of the predicted class concerning the activations of the final neural network layer to produce the heatmap, thereby offering class-specific visualizations. 

\subsection{Extending Grad-CAM to Videos}
\indent A straightforward approach to adapting feature attribution methods from images to videos is to consider each frame as an individual image. However, this bears a significant limitation: it neglects the temporal interplay among frames, a crucial aspect for understanding the actions in videos.\\
\indent This limitation can be overcome by utilizing a collection of frames as input, thus preserving the temporal connections.  These frames are input to the Video Swin Transformer model \cite{https://doi.org/10.48550/arxiv.2106.13230}, which was pre-trained on the Kinetics-400 dataset. 
% The transformer blocks in the Video Swin Transformer are capable of capturing both spatial and temporal features of the input by jointly analyzing multiple frames.  Extending the methodology akin to Grad-CAM's in the context of image classification, we broaden the concept to videos by analyzing the output of the final layer. 
Subsequently, gradients are computed with respect to the corresponding activation maps. Illustrative instances displaying Grad-CAM outputs for inputs related to the class \emph{playing tennis} for the Swin Transformer model are presented in Figure ~\ref{fig:gradcam_tennis_1}. The overlaid heatmap demonstrates that the outputs of the Grad-CAM are concentrated in the moving regions associated with the tennis racket.

% \begin{figure}[ht]
%   \centering
  
%   \includegraphics[width=0.45\textwidth]{figs/gradCAM_video.png}
%     \label{fig:original}
    
% \captionsetup{font=small}
%   \caption{Extending Grad-CAM to videos using 3D CNNs.}
%   \vspace{-1em}
% \end{figure}\label{fig:gradcam_video}

% \begin{figure}[ht]
%     \vspace{-2mm}
%     \centering
%  	\includegraphics[width=1.0\linewidth]{figs/gradCAM_tennis_smaller.png}
%  	\captionsetup{font=small}
%  	\caption{Outputs of Grad-CAM with respect to the class \textit{playing tennis} for the video recognition model I3D. We can clearly see from the heatmap that the region of movement for the tennis racquet is highlighted. }
%  	\label{fig:gradcam_tennis}
% \end{figure}

% \subsection{Limitations of Grad-CAM for videos}
\subsection{Limitations in Feature Attribution Methods}
Extending feature attribution methods designed for image recognition models to work well with videos is a non-trivial task and raises concerns regarding robustness. A significant bottleneck lies in processing temporal data in the correct semantic order. For example, the same set of frames played in reverse order should mean the output should be reversed for video action recognition \textit{i.e.,} a person \textit{picking up} an object and \textit{placing down} an object would exhibit identical spatial data, but opposite temporal directionality. While coarse details suffice for overall action labeling in videos, they may not meet the requirements of explaining complex models such as the Video Swin Transformer. To ensure reliable and smooth label explanations across frames, a finer time domain analysis becomes necessary. Background elements further complicate Grad-CAM decisions, as seen in Figure \ref{fig:gradcam_limitation_tennis}, where the focus shifts to the tennis court rather than the players. Further, Grad-CAM remains a local explanation method, implying that the analysis needs to be done individually for each attribution to draw any class-level conclusions. While this process is feasible for images, which can be viewed in batches, it becomes impractical for videos.

%Due to these limira in image feature attribution methods, interest has grown in more meaningful methods tailored for human-level understanding, such as 
% concept attribution, exemplified by 
%Testing with Concept Activation Vectors (TCAV)~\cite{kim2018interpretability}, which we extend to work with videos in Section~\ref{sec:videotcav}.
% We delve into this further and propose novel approaches to extending this idea to videos in Section \ref{sec:videotcav}.

% Additionally, as mentioned earlier, these attribution maps suffer from significant issues such as \textit{ghost attributions}, where invisible objects are highlighted, and \textit{continuity problems}, where highlighted objects flicker, or attribution areas flip between frames. 
Beyond these domain-specific issues, attribution methods exhibit several other weaknesses. In \cite{adebayo2018local}, the authors demonstrated that networks with random weights generate similar attribution maps as trained networks. Additionally, attribution methods are vulnerable to adversarial attacks \cite{ghorbani2019interpretation}, changes in data preprocessing \cite{kindermans2019reliability}, and even the introduction of random noise \cite{DBLP:journals/corr/abs-1806-08049}. Their widespread use poses particular concerns in high-risk applications such as medical imaging \cite{arun2021assessing}. Due to these limitations in feature attribution methods, interest has grown in more meaningful methods tailored for human understanding, such as concept attribution, introduced in Testing with Concept Activation Vectors (TCAV) ~\cite{kim2018interpretability}, that we extend to videos in Section~\ref{sec:videotcav}.

\section{Video TCAV}
\label{sec:videotcav}
\subsection{TCAV Basics}
\begin{figure}[htbp]
    \centering
    \includegraphics[width=0.80\linewidth]{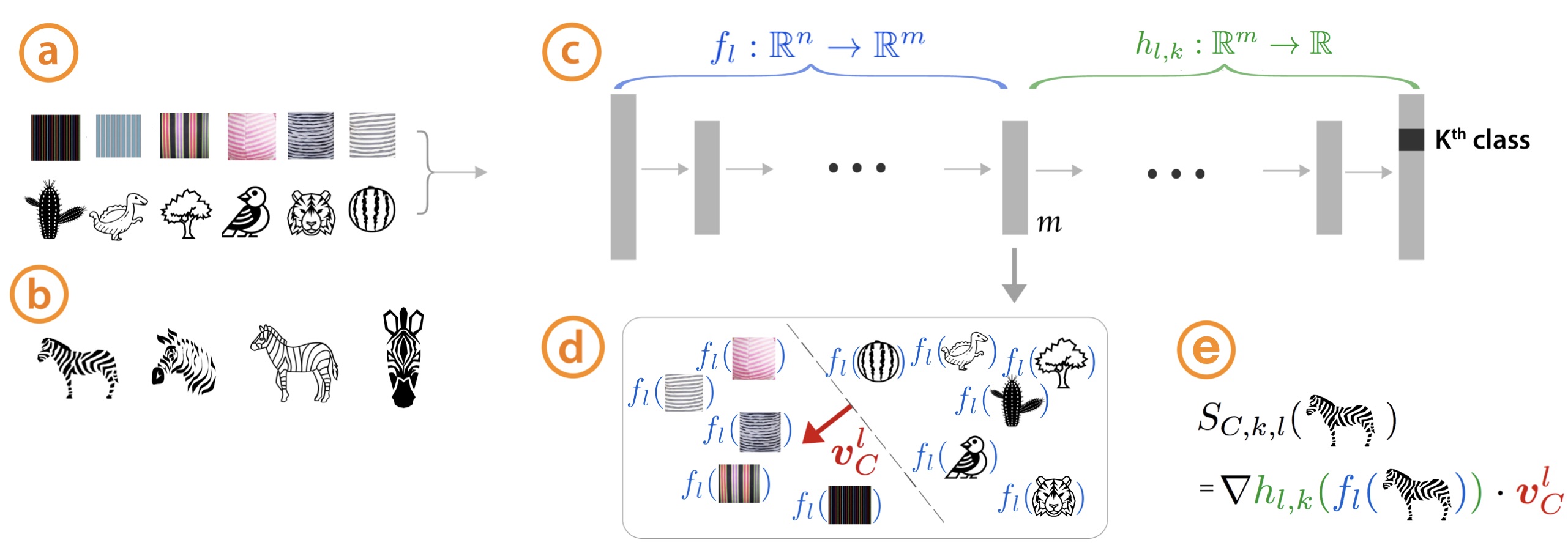}
    \caption{TCAV process in Image Classification. Image taken from \cite{kim2018interpretability}. Best viewed zoomed. }
    \label{tcav}
\vspace{-1em}
\end{figure}

TCAV offers a popular alternative to feature attribution methods like Grad-CAM. Unlike granular, pixel-level changes, TCAV focuses on abstract details. It is widely accepted that the human brain converts raw pixel values into high-level concepts like texture, specific objects, and their interactions \cite{Epstein2019}. Similarly, neural networks encode analogous high-level concepts in the output of embedding layers. TCAV attempts to derive explanations by analyzing properties within these embedding spaces. \\ 
\indent In TCAV, a concept is delineated by a collection of inputs that characterize it. For instance in image classification, the concept of \emph{stripedness} can be delineated by a set of images portraying striped objects (\Circled{a} in Figure \ref{tcav}). When mapped to the corresponding feature space, the representations of this collection would ideally be aligned with a specific \emph{stripedness} axis (\Circled{c} and \Circled{d} in Figure \ref{tcav}). On the other hand, a collection of random images would not display any such alignment. Thus, if we were to find a hyperplane that separates the embeddings of a set of concept-specific images and a set of random (or control) images, the normal to that hyperplane should give us a representation of the concept. This normal may also be considered a basis vector of the feature space associated with that concept (e.g., a degree of \emph{stripedness}), and is called the Concept Activation Vector (CAV). Now, given any input for our specific task, say a picture of a zebra (\Circled{b} in Figure \ref{tcav}), the directional derivative of its embedding along the CAV would measure the sensitivity of the embedding to the concept. Figure \ref{tcav} taken from \cite{kim2018interpretability} shows a graphical representation of the entire process discussed above. 
This scalar quantity may be aggregated differently for multiple inputs to get a more robust value. Moreover, since any learned hyperplane will return a CAV, even if it does not separate the data well, this process is usually repeated for multiple control sets to ensure the sensitivity is statistically significant. Thus, TCAV allows us to quantify the importance of any
given concept to a specific step in the neural networks. It is
especially useful for comparing concepts, which is
hard with attribution maps. 
\subsection{Video-TCAV Components}
\begin{figure}[ht]
    \vspace{-2mm}
    \centering
 	\includegraphics[width=0.85\linewidth]{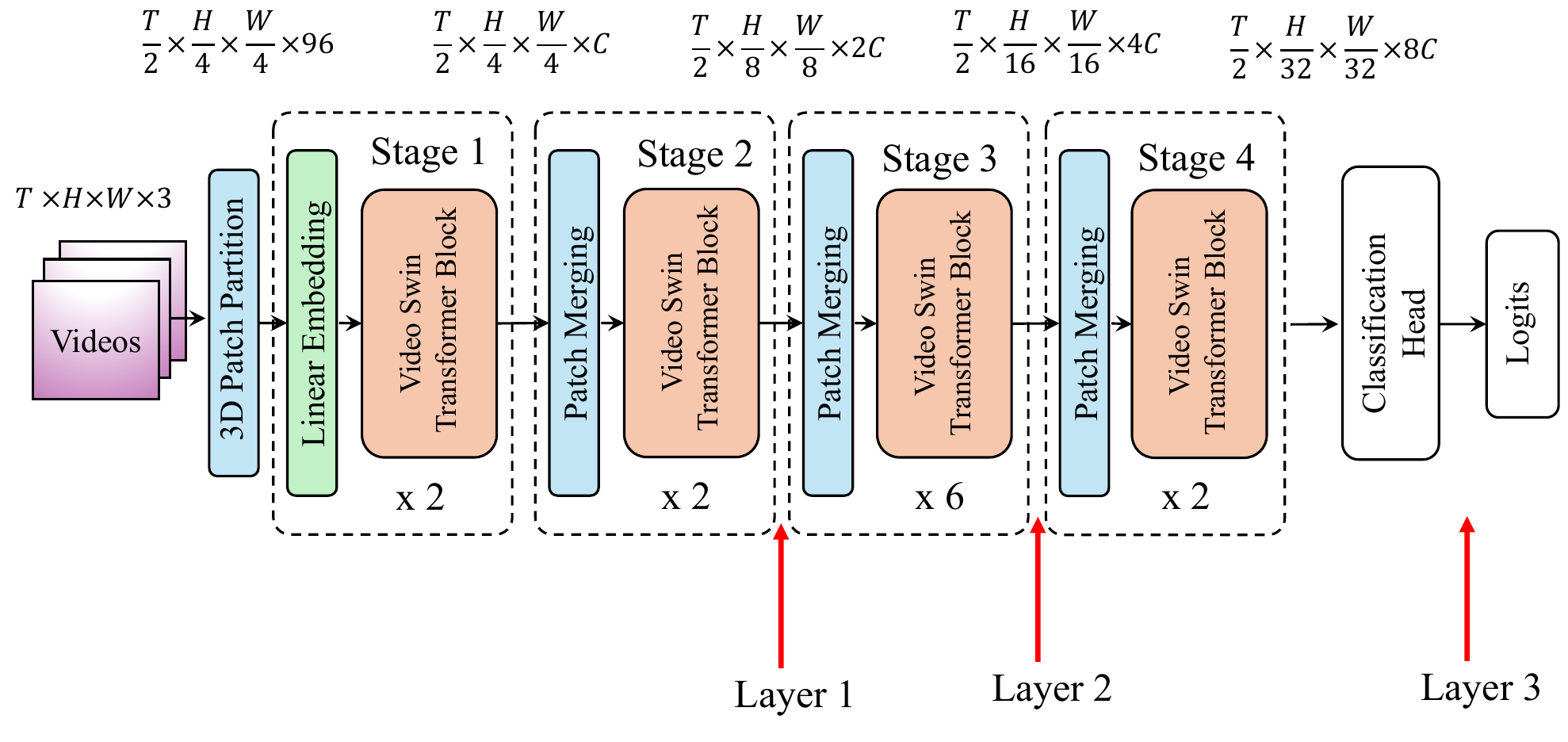}
 	\captionsetup{font=small}
 	\caption{Video Swin Transformer block diagram. The 3 layers with red arrows whose activations we study while testing CAVs are marked. Best viewed zoomed. }
 	\label{fig:swin}
\vspace{-0.5em}
\end{figure}

This section discusses the various components of the proposed Video-TCAV framework.  \\ \\
\textbf{Video Action Recognition Model:} Similar to TCAV, Video-TCAV is a post-training explanation method, requiring a pre-trained Video Action Recognition model. We opt for the state-of-the-art Video Swin Transformer model trained on the Kinetics-400 dataset for quantitative evaluation of post-training explanations. Figure \ref{fig:swin} illustrates the schematic diagram of the Video Swin Transformer, including the three layers whose activations we utilize in testing CAVs in Video-TCAV.  \\ \\
\textbf{Concepts:} The most important component of Video-TCAV, similar to TCAV, is the concepts we want to test. In contrast to image classification, where selecting concepts is relatively simple, generating concepts for video classification poses a more intricate challenge. This complexity arises because choosing clips to represent concepts for Video-TCAV, similar to image crops in the case of TCAV, is not trivial. We generate two categories of concepts: \begin{itemize}
    \item \emph{Spatial Concepts}: These are simply images of objects repeated temporally to form videos. This is done as the Video Swin Transformer can only take video input. 
    \item \emph{Spatiotemporal Concepts}: These are objects or humans in motion. These are obtained using YOLO-v7 \cite{https://doi.org/10.48550/arxiv.2207.02696} object detector frame-wise to track a concept of our choice.  
\end{itemize}

\subsection{Generating Video-TCAV Concepts}
\begin{figure}[ht]
    \vspace{-2mm}
    \centering
 	\includegraphics[width=0.75\linewidth]{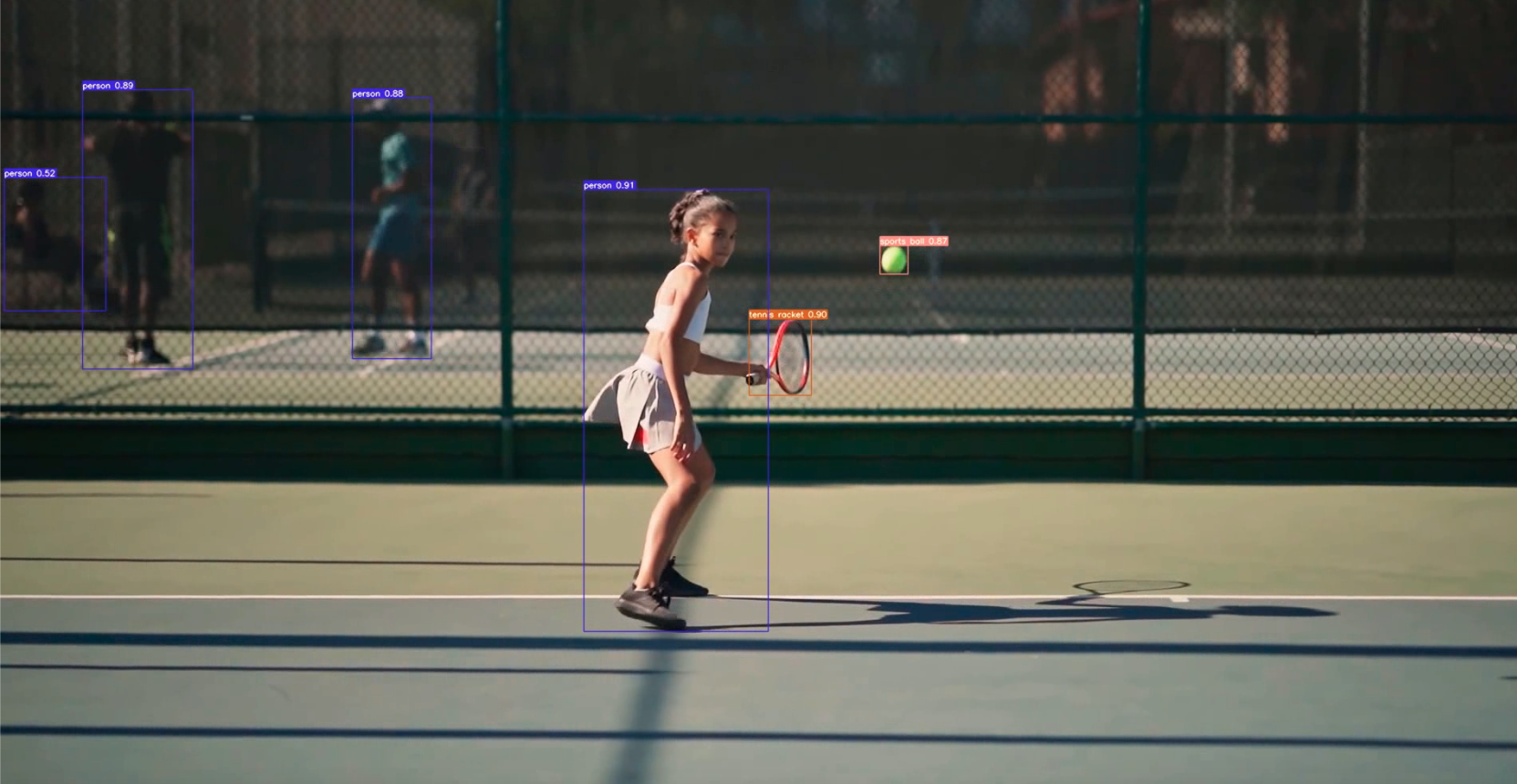}
 	\captionsetup{font=small}
 	\caption{YOLO-v7 detections on a frame of a video from the \emph{playing tennis} class. Best viewed zoomed.}
 	\label{fig:yolo}
\end{figure}

\begin{figure}[ht]
    \vspace{-2mm}
    \centering
 	\includegraphics[width=0.8\linewidth]{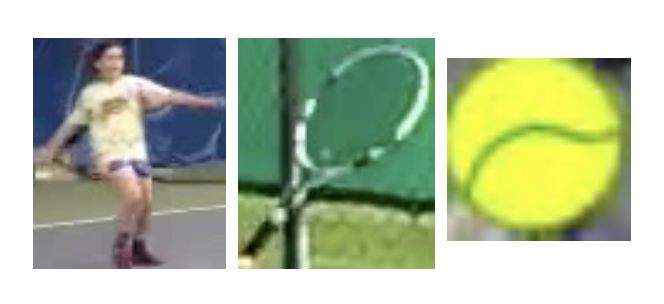}
 	\captionsetup{font=small}
 	\caption{Exemplars of Spatial Concepts: person playing tennis, tennis racket, and sports ball generated from Kinetics-400 dataset. }
 	\label{fig:spatialconcepts}
\end{figure}

\begin{figure}[ht]
    \vspace{-2mm}
    \centering
 	\includegraphics[width=0.70\linewidth]{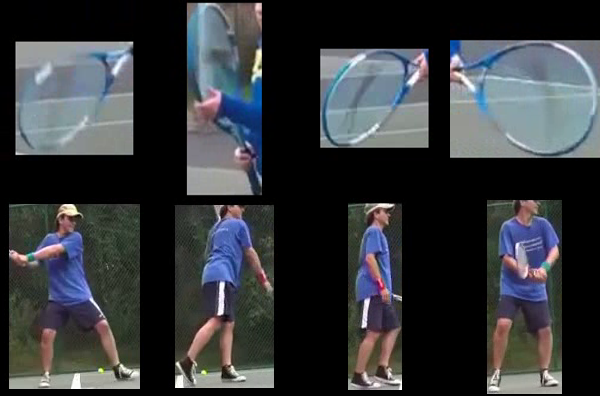}
 	\captionsetup{font=small}
 	\caption{Exemplars of Spatiotemporal Concepts: movement of tennis racket in a match and person playing tennis. We show some sampled frames across each concept video.}
 	\label{fig:temporalconcepts}
\end{figure}

We adopt a machine-assisted method to generate spatial and spatiotemporal concepts for Video-TCAV. Utilizing YOLO-v7, we detect objects in videos and produce spatial and spatiotemporal crops for concept testing in Video-TCAV. Given that the object detector is not flawless, we manually verify all concept video crops to ensure accuracy in our experiments. Following \cite{kim2018interpretability}, we present our visualization and results for a single action class \emph{playing tennis} for visualization and results demonstration. Figure \ref{fig:yolo} illustrates sample detections generated by YOLO-v7 on a video frame from the \emph{playing tennis
 class}. Notably, YOLO-v7 provides accurate detections of a tennis racket, sports ball, and individuals, which we consider potential concepts aiding in explainability of videos belonging to \emph{playing tennis} class. \\ \\ %More details on concept generation can be found in the Section \ref{sec:addition} of Supplementary material.
 \noindent \textbf{Generating Spatial Concepts:} We execute the YOLO-v7 object detector on all videos from concept generating split of the test set videos of the Kinetics-400 dataset labeled as \emph{playing tennis'}. Since YOLO-v7 is trained on object detection with 80 different object classes, it leads to multiple object detections in each video frame. For our experiments, we considered all the per-frame detections from the three classes, i.e., person, tennis racket, and sports ball, as concepts that we want to test, and all the other detections are used to generate random concept sets. Figure \ref{fig:spatialconcepts} illustrates our generated spatial concepts. The generated spatial concepts are temporally repeated to generate a static video of 3-minute duration. \\ \\
\noindent \textbf{Generating Spatiotemporal Concepts:} Contrary to spatial concepts, generating spatiotemporal concepts is non-trivial. The basic idea of generating spatiotemporal concepts is tracking a particular instance of an object across frames, like the person playing tennis, a tennis racket, or a sports ball across video frames. This is non-trivial as there might be multiple instances of the same object class in the frame, and YOLO-v7 results in multiple detections and it is necessary to choose the same instance of the object in consecutive frames to prevent the concept video from changing in terms of content. We ensured only one instance of the object was tracked across frames, and we took a spatiotemporal crop based on the detections returned by YOLO-v7. Since the size of the detections of a particular object changes across frames, we generate the concept video with the largest-sized object detected in consecutive frames and pad the other smaller-sized frames, positioning them at the center. Figure \ref{fig:temporalconcepts} illustrates our generated spatiotemporal concepts.

%updating it holdon

\vspace{-5mm}
\subsection{Video-TCAV Experiments}
\textbf{Experimental Setup:} We replicate the experiments of \cite{kim2018interpretability} to demonstrate the advantages of our pipeline. We collect 30 videos that have the ground truth label as \emph{playing tennis} and generate 25-30 videos per concept. We use these videos and concepts in our Video-TCAV framework. For random concepts, we randomly select 30 videos from our entire corpus of concepts, including those not related to playing tennis e.g. dining and dancing. All experiments use the relative sign-count variant of TCAV. In this, we consider a group of multiple concepts at a time, and learn a one-vs-rest classifier for each concept, which gives us the CAV for that concept. The fraction of data points positively affected by the CAV is the relative TCAV for that concept. Using relative TCAVs helps disentangle the effect of correlation between different concepts. A more detailed discussion may be found in \cite{kim2018interpretability}.
\subsection{Results \& Discussion}

\begin{figure}[htbp]
    \centering
    \includegraphics[width=0.80\linewidth]{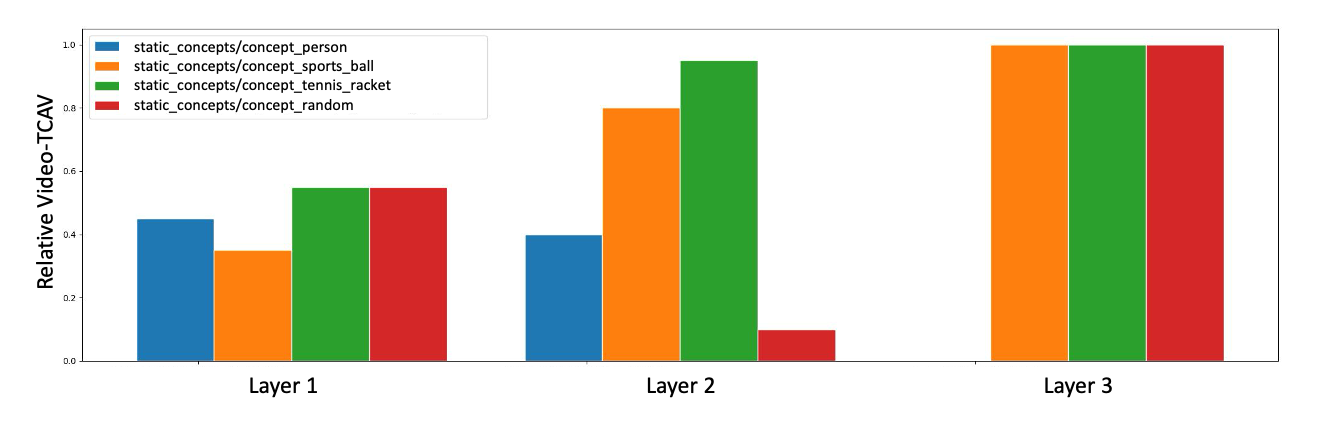}
    \caption{Relative TCAVs for static concepts. }
    \label{fig:spatial}
\vspace{-1em}
\end{figure}

\textbf{Spatial Concepts:} The generated spatial concepts of static frames with no movement information serve as our baseline. Figure \ref{fig:spatial} shows the relative TCAVs of the tested static concepts in different layers. While all concepts show some relative importance, it is hard to distinguish their importance from a random concept, especially in the initial layers. This is in line with our expectations, as static concepts should provide some information the context of the video (e.g. the presence of a tennis racket indicates tennis being played) but should not be able to pinpoint the exact action performed. Another aspect to note is that the importance of the random concept persists until the last layer, which validates the static features are not fully discriminative for the action recognition task.

\begin{figure}[htbp]
    \centering
    \includegraphics[width=0.80\linewidth]{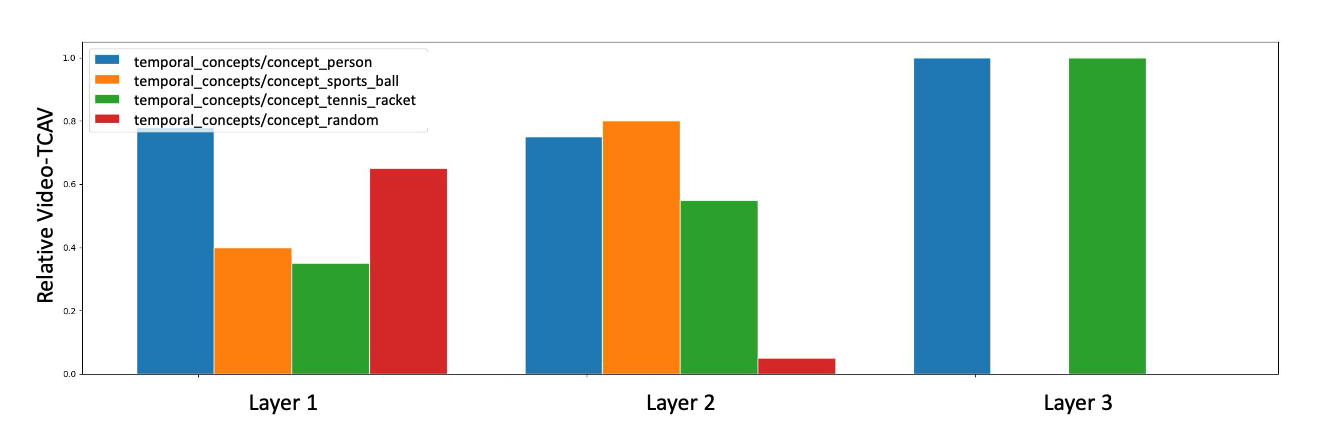}
    \caption{Relative TCAVs for dynamic concepts}
    \label{fig:dynamic}
\vspace{-1em}
\end{figure}

\noindent\textbf{Spatio-Temporal Concepts:} The dynamic spatio-temporal concepts, on the other hand, have more predictive power than the static concepts. Figure \ref{fig:dynamic} shows the relative TCAVs of the tested dynamic concepts in different layers. In the initial layers, the importance of the dynamic concepts is more distinguishable from random, but is overall similar to static case, which is in line with the understanding that initial layers extract general-purpose features. However, in later layers, the importance of temporal concepts is dramatically more pronounced. In contrast with the static concepts, the random concept holds negligible importance in the last layer. Interestingly, this increase of importance of temporal information with depth correlates well with how the human brain perceives motion, with visual features first processed by the V1 area, then motion being handled later in the V5/MT area \cite{10.1093/cercor/3.2.79}.

\begin{figure}[htbp]
    \centering
    \includegraphics[width=0.80\linewidth]{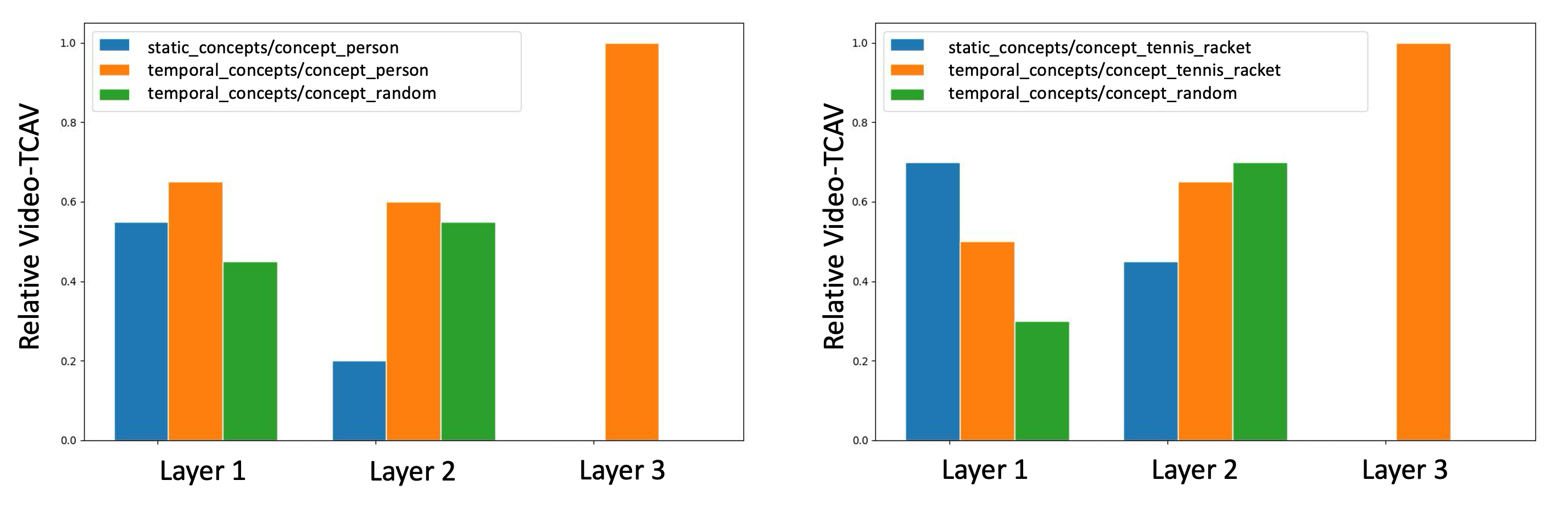}
    \caption{Relative TCAVs for static vs dynamic concepts}
    \label{fig:static_v_dynamic}
\vspace{-0.75em}
\end{figure}

\noindent\textbf{Spatial vs Spatio-Temporal Concepts:} Next, we directly compare the spatial and spatio-temporal versions of a specific concept. Figure \ref{fig:static_v_dynamic} shows their relative TCAVs in different layers. Evidently, when compared side-by-side, the temporal concepts dominate over static concepts in most layers of the network, with this effect becoming more pronounced with depth. It is particularly interesting to note how the final layer exclusively prefers the temporal concept for prediction compared to the static version of the concept. \\

\begin{figure}
    \centering
    \includegraphics[width=0.80\linewidth]{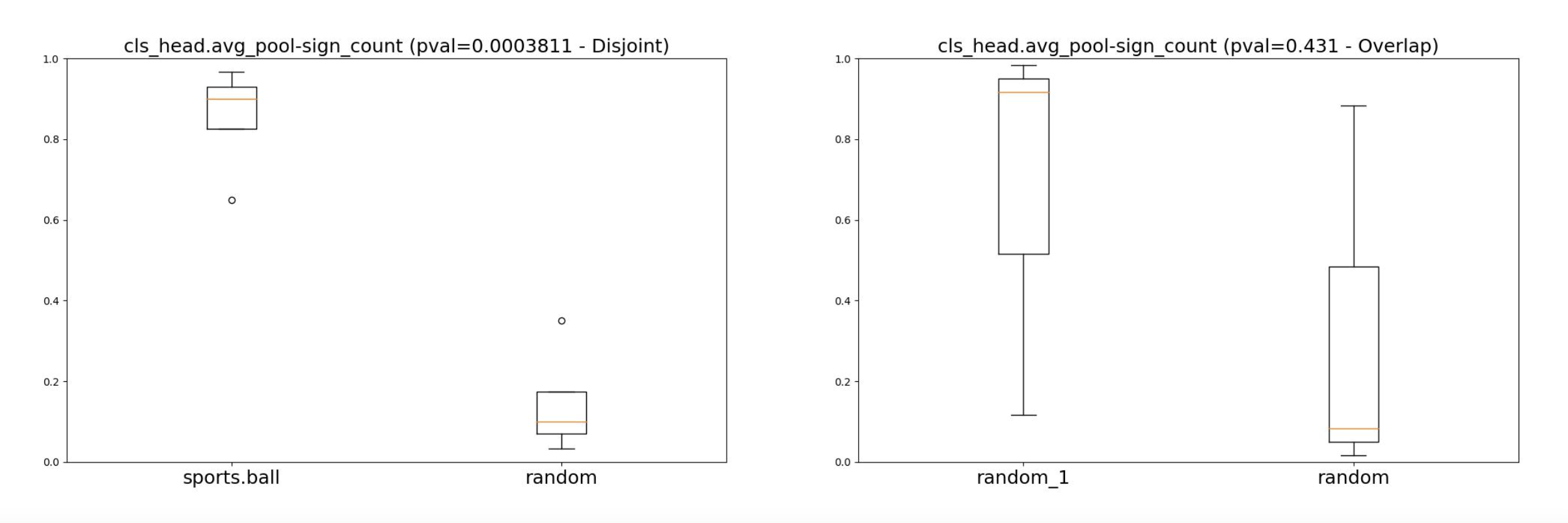}
    \caption{Results of hypothesis testing with spatiotemporal concepts. Best viewed zoomed.}
    \label{fig:pval}
\end{figure}

\noindent \textbf{Statistical Testing:} We tested the validity of the scores with 10 other random sets and applied a two-sided t-test with Bonferroni correction, as in \cite{kim2018interpretability}. The results are in the Figure \ref{fig:pval}. The $p < 0.05$ for our concepts implies that the CAVs we found are robust and statistically significant, where as the CAVs learnt by separating random concepts are not meaningful ($p > 0.05$).

\section{Conclusion and Future Work }
\label{sec:conclude}
In this work, we proposed Video-TCAV, a new approach for generating human-friendly concepts and quantitatively measuring their impact on the decision process of deep networks used for action recognition. 
Further, we provide preliminary results on the Video Swin Transformer to demonstrate the success and pitfalls of our method. 
However, more experiments using other video action recognition models would strengthen the effectiveness of Video-TCAV. 
Additionally, Video-TCAVs may also share some of
the weaknesses of attribution methods, such as~\cite{DBLP:journals/corr/abs-2110-07120}, which demonstrates adversarial attacks on TCAV that can artificially raise or lower the importance of a particular concept, leading to bizarre conclusions.
Generating concepts for Video-TCAV is challenging. In this work, we proposed a method using YOLO-v7 object detector. An alternative research direction could involve utilizing text-to-video diffusion models, as  in \cite{stablevideodiffusion}, to generate concept clips based on human-interpretable text prompts. We hope that this work serves as a starting direction for developing scalable explainable methods for Video Action Recognition. 
%Another potential solution is language-guided spatiotemporal video cropping similar to \cite{narasimhan2021clipit}, which should offer a method to generate video clips that represent some human interpretable concept. 
{
    \small
    \bibliographystyle{ieeenat_fullname}
    \bibliography{main}
}


\end{document}